\documentclass[letterpaper, 10 pt, conference]{ieeeconf}  

\IEEEoverridecommandlockouts                              

\title{\LARGE \bf
Designing Robots with, not for: A Co-Design Framework for Empowering Interactions in Forensic Psychiatry}

\author{Qiaoqiao Ren, Remko Proesmans, Arend Pissens, Lara Dehandschutter, \\
William Denecker, Lotte Rouckhout, Joke Carrette, Peter Vanhopplinus, Tony Belpaeme and Francis wyffels
\thanks{Qiaoqiao Ren, Remko Proesmans, Tony Belpaeme and Francis wyffels are with AIRO-IDLab, Department of Electronics and Information Systems, Ghent University -- imec; Joke Carrette are Peter Vanhopplinus with the Forensic Psychiatric Center Ghent
{\tt\small Qiaoqiao.Ren@ugent.be}
}
}

\usepackage{graphicx}
\usepackage{subcaption}
\usepackage{color}
\usepackage{textcomp}
\usepackage{csquotes}
\usepackage{hyperref}

\usepackage{caption} 
\usepackage{tabularx}
\usepackage{float}
\usepackage{array}  
\usepackage{booktabs} 
\usepackage{multirow}
\usepackage{makecell}
\usepackage{arydshln} 
\usepackage{todonotes}
\usepackage{eurosym}
\usepackage{balance}

\usepackage{siunitx}
\usepackage{cite}
\begin{document}


\maketitle

\begin{abstract}

Forensic mental health care involves the treatment of individuals with severe mental disorders who have committed violent offences. These settings are often characterized by high levels of bureaucracy, risk avoidance, and restricted autonomy. Patients frequently experience a profound loss of control over their lives, leading to heightened psychological stress—sometimes resulting in isolation as a safety measure. In this study, we explore how co-design can be used to collaboratively develop a companion robot that helps monitor and regulate stress while maintaining tracking of the patients' interaction behaviours for long-term intervention. We conducted four co-design workshops in a forensic psychiatric clinic with patients, caregivers, and therapists. Our process began with the presentation of an initial speculative prototype to therapists, enabling reflection on shared concerns, ethical risks, and desirable features. This was followed by a creative ideation session with patients, a third workshop focused on defining desired functions and emotional responses, and we are planning a final prototype demo to gather direct patient feedback. Our findings emphasize the importance of empowering patients in the design process and adapting proposals based on their current emotional state. The goal was to empower the patient in the design process and ensure each patient's voice was heard.


\end{abstract}

\section{Introduction}

Forensic mental health care refers to the treatment of individuals who have committed violent offenses and are diagnosed with severe mental disorders \cite{mullen2000forensic}. Most forensic psychiatric patients are treated in low‑ or medium‑security services; only a small minority require the stringent measures of a high‑security ward. Some participants fall within this higher‑risk subgroup, making intensive supervision essential. In these units, therapy concentrates on practising risk management in a closely monitored yet supportive setting. By helping patients recognise, tolerate, and regulate their emotions—together with any associated maladaptive behaviours—clinicians create the conditions for meaningful progress before patients transition to lower‑security facilities.


Socially intelligent robots have been introduced in various healthcare contexts to support emotional well-being and behaviour monitoring \cite{scoglio2019use}. In addition, previous research has shown that tactile interaction with a robot can reduce stress levels \cite{ren2024tactile} and that robots can decode emotions from touch input \cite{ren2024conveying}. However, in forensic psychiatry, the application of such technologies remains underexplored—particularly from a perspective that includes the lived experiences and needs of patients themselves. Most existing robotic interventions prioritize compliance and behavioural correction, rather than relational sensitivity, user agency, or contextual nuance.

\begin{figure*}
    \centering
        \includegraphics [height=6cm]{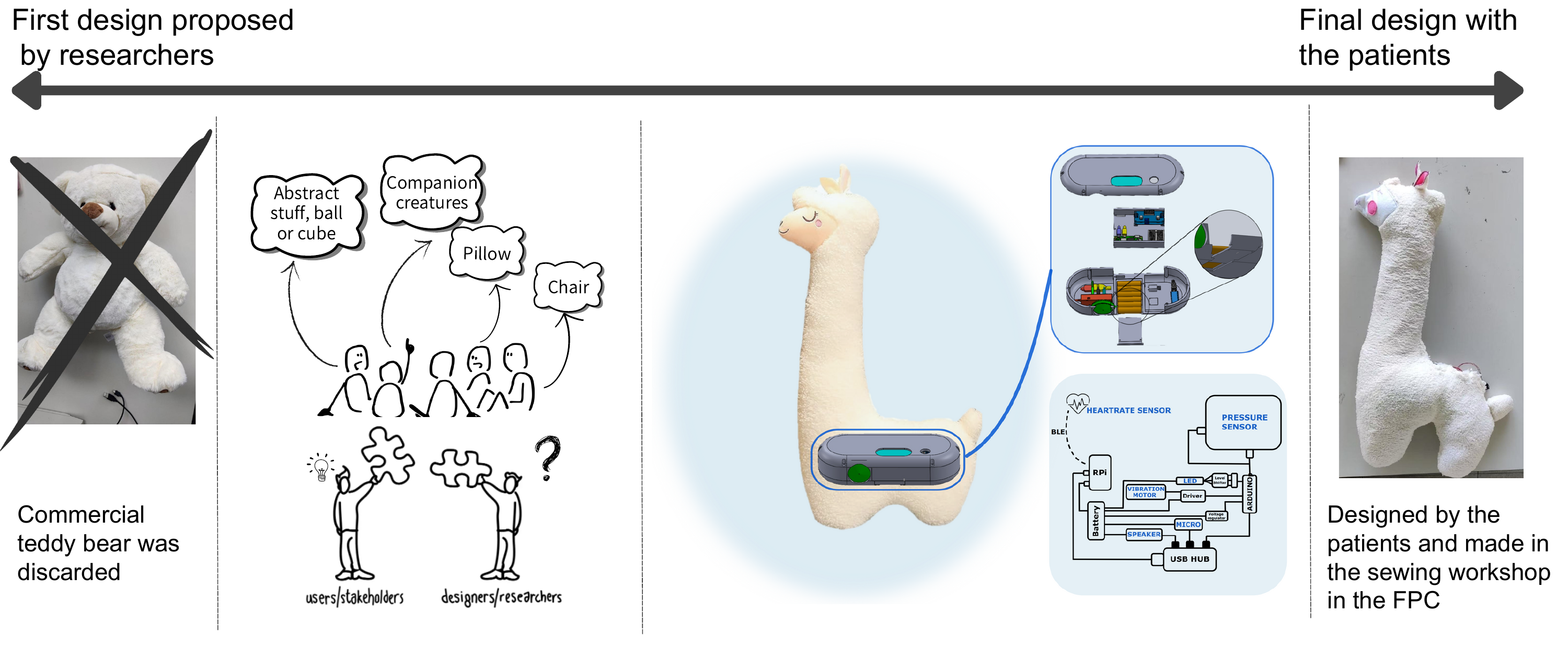}
    \caption{Co-design process of the companion robot. After the therapists rejected the initial concept, patients proposed different directions, such as abstract forms (e.g., ball/cube) or animal-like shapes (e.g., cat). They eventually co-designed an alpaca-inspired prototype. A modular electronic system was developed to support future customization.}
    \label{fig::sam}
\end{figure*}

As shown in the Fig.~\ref{fig::sam}, we explore a co-design project on how robotic \textit{companion creatures} could support emotional well-being in a forensic psychiatric clinic. Through a series of four workshops involving patients, caregivers, and therapists, we developed a speculative framework for designing emotionally responsive and empowering robotic systems tailored to the needs of forensic care settings.

\section{Methods}
\subsection{Context and Participants}
This study was in collaboration with a forensic psychiatric centre (FPC). The co-design process involved a total of four workshops, with higher‑risk subgroup patients, therapists and technical support. Each workshop was facilitated by researchers with experience in human-robot interaction, participatory design, and mental health.

\subsection{Workshop Design}
Each workshop had a specific focus and built upon the outcomes of the previous sessions:
\begin{itemize}
    \item \textbf{Workshop 1: Therapist-Oriented Exploration}\\
    We presented a teddy bear equipped with an air-pressure sensor and a microphone as an example to therapists to initiate discussions about ethical concerns, potential risks, and shared values in the clinical context.
    
    \item \textbf{Workshop 2: Patient-Led Brainstorming}\\ Patients shared their ideas for a companion creature, suggesting concepts such as abstract forms, cushions, or animal-like robots. They also expressed a desire to be informed about their own stress levels and whether they had successfully regulated their emotions. Additionally, they mentioned a preference for receiving reward-based feedback when they managed to do so.
    
    \item \textbf{Workshop 3: Functionality and Form Exploration} \\
    Participants helped define desired features, emotional responses, and physical forms of the robot.
    
    \item \textbf{Workshop 4: Tactile Interaction Design}\\
    Modular interactive patterns,  reactive behaviours and use cases were co-developed based on patient input. Patients designed and fabricated the skin of the companion creature prototype by themselves.
   
\end{itemize}

\subsection{Data Collection and Analysis}
Workshops were documented through field notes. Insights were iteratively validated with the therapist and patient to ensure alignment with patient care goals. All the related materials we used for the project have been reported and registered to the forensic psychiatric centre.

\section{Proposed framework}





\subsection{A Co-Design Framework for Forensic Companion Robots}
Based on the four workshops and thematic analysis, inspired by reflective transformative design \cite{hummels2009reflective} and aspects of participatory design \cite{spinuzzi2005methodology}, we propose a preliminary co-design framework for developing emotionally responsive robots in forensic psychiatric settings. The framework comprises the following interdependent phases and design commitments:

\begin{enumerate}
    \item \textbf{Context Grounding and Stakeholder Alignment}:
    Initiate design by collaboratively identifying care constraints, ethical boundaries, and shared goals with clinical staff and therapists. Use speculative prototypes to surface unspoken tensions or risks.

    \item \textbf{Patient-Centered Ideation}:
    Facilitate creative, emotionally safe sessions for patients to imagine companion roles, preferred sensory interactions, and meaningful physical forms. Prioritize emotional expression over functional output at this stage.

    \item \textbf{Iterative Prototyping with Material Constraints}:
    Translate patient ideas into modular, tangible prototypes while acknowledging limitations of time, safety, and materials. Encourage adaptation and refinement with each cycle of feedback.

    \item \textbf{Situated Interaction Testing}:
    Evaluate prototypes in situ with patients in a familiar clinical setting. Emphasize emotional comfort, perceived control, and non-verbal feedback.

    \item \textbf{Blended Integration Planning}:
    Ensure the robot's role complements—rather than replaces—human caregivers. Design affordances that can flexibly support both autonomous use and therapist-guided interactions.

\end{enumerate}


\section{Discussion and Design Implications}

Our findings highlight several key considerations for designing socially intelligent robots in forensic psychiatric settings. First, empowerment must be a central design goal~\cite{donoso2014increasing}. Patients should have meaningful choices and active roles in shaping the robot’s form and function. For example, while the alpaca doll designed by patients may resemble a typical teddy bear—and could similarly be perceived as childish—its true value lies in the fact that it was co-created. This reflects the IKEA effect, a psychological phenomenon where individuals attach greater value to items they’ve had a hand in making, regardless of the final product’s quality \cite{norton2012ikea}. Simply giving a generic toy to a patient might be unhelpful—or even distressing—but engaging them in the design process promotes a sense of ownership, personal dignity, and emotional connection. Even when material constraints limit the final prototype, the act of choosing and crafting remains empowering. Second, contextual adaptability is essential. Emotional states in psychiatric care environments are highly dynamic. Companion robots must be able to respond to moment-to-moment shifts in mood, stress, and interpersonal interactions. Static or overly scripted behaviours may not only lose their relevance but could also have detrimental effects. Third, designing for relational trust is critical. The robot’s physical form, behaviour, and presence must avoid evoking feelings of surveillance, control, or infantilization. In clinical contexts where trust can be fragile, the robot should be perceived as a supportive companion rather than a disciplinary figure. While the robot should empower patients, it must also support caregivers. Designers need to strike a balance between respecting clinical workflows and enabling responsive, patient-centred interaction. This dual perspective is crucial for building therapeutic alliances and ensuring ethical integration into care practices.

\section{Limitations and Future Work}
As an explorative study, our work involved four co-design workshops with a limited number of patients, constrained by ethical considerations. While these sessions provided insights, scaling up the approach remains a key challenge. Future research will focus on exploring how such co-creation processes might inspire new tools or solutions within psychiatric care.

\bibliographystyle{ieeetr}
\bibliography{references}

\vfill

\end{document}